\documentclass[letterpaper, 10 pt, conference]{ieeeconf}  

\IEEEoverridecommandlockouts                              

\usepackage{epsfig}
\usepackage{amsmath}
\usepackage{color}
\usepackage{graphicx}
\usepackage{cleveref}
\DeclareTextCommandDefault{\textcopyright}{\textcircled{R}}

\title{\LARGE \bf
Towards Automatic 3D Shape Instantiation for Deployed Stent Grafts: 2D Multiple-class and Class-imbalance Marker Segmentation with Equally-weighted Focal U-Net
}

\author{Xiao-Yun~Zhou$^{1}$, Celia~Riga$^{2}$, Su-Lin~Lee$^{1}$ and Guang-Zhong~Yang$^{1}$
\thanks{$^{1}$Xiao-Yun Zhou, Su-Lin Lee and Guang-Zhong Yang are with the Hamlyn Centre for Robotic Surgery, Imperial College London, UK {\tt\small xiaoyun.zhou14@imperial.ac.uk}}
\thanks{$^{2}$Celia Riga is with the Regional Vascular Unit, St Mary's Hospital, London, UK and the Academic Division of Surgery, Imperial College London, UK}
}

\begin{document}
\maketitle
\thispagestyle{empty}
\pagestyle{empty}

\begin{abstract}
Robot-assisted Fenestrated Endovascular Aortic Repair (FEVAR) is currently navigated by 2D fluoroscopy which is insufficiently informative. Previously, a semi-automatic 3D shape instantiation method was developed to instantiate the 3D shape of a main, deployed, and fenestrated stent graft from a single fluoroscopy projection in real-time, which could help 3D FEVAR navigation and robotic path planning. This proposed semi-automatic method was based on the Robust Perspective-5-Point (RP5P) method, graft gap interpolation and semi-automatic multiple-class marker center determination. In this paper, an automatic 3D shape instantiation could be achieved by automatic multiple-class marker segmentation and hence automatic multiple-class marker center determination. Firstly, the markers were designed into five different shapes. Then, Equally-weighted Focal U-Net was proposed to segment the fluoroscopy projections of customized markers into five classes and hence to determine the marker centers. The proposed Equally-weighted Focal U-Net utilized U-Net as the network architecture, equally-weighted loss function for initial marker segmentation, and then equally-weighted focal loss function for improving the initial marker segmentation. This proposed network outperformed traditional Weighted U-Net on the class-imbalance segmentation in this paper with reducing one hyper-parameter - the weight. An overall mean Intersection over Union (mIoU) of $0.6943$ was achieved on $78$ testing images, where $81.01\%$ markers were segmented with a center position error $<1.6mm$. Comparable accuracy of 3D shape instantiation was also achieved and stated. The data, trained models and TensorFlow codes are available on-line.
\end{abstract}

\section{Introduction}
\begin{figure}[t]
\centering
\includegraphics[width=\linewidth]{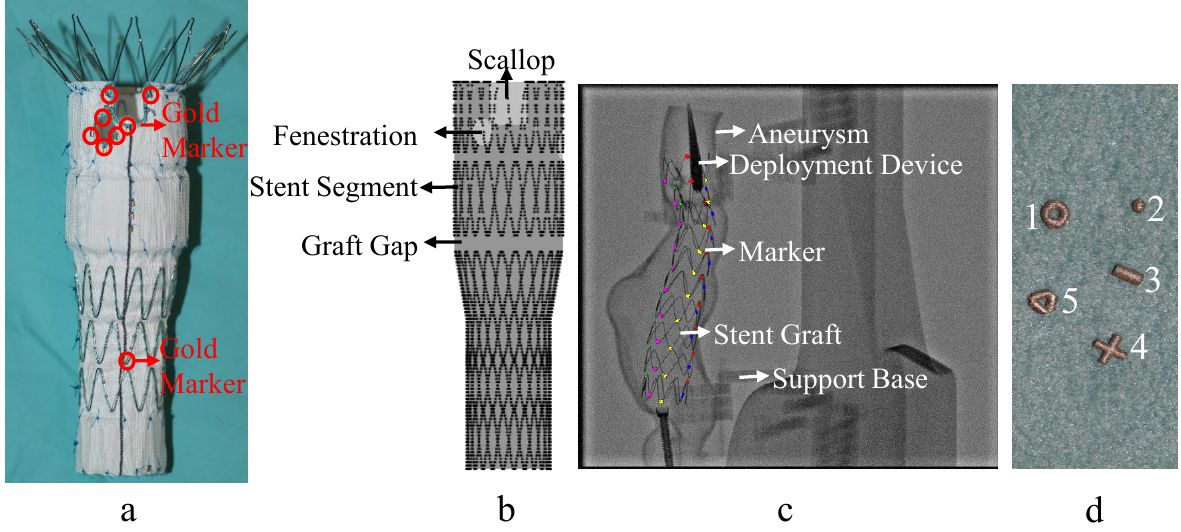}
\caption{(a) A real fenestrated stent graft with commercial gold markers indicating the fenestration and the scallop; (b) the mathematical model of the real fenestrated stent graft; (c) an experimental fluoroscopy image example with five markers - the red, green, blue, yellow, and purple color indicate marker 1, marker 2, marker 3, marker 4, and marker 5 respectively, this marker sequencing is valid across the whole paper; (d) 3D printed customized markers.}
\label{fig: introduction}
\end{figure}

Abdominal Aortic Aneurysm (AAA) is an enlargement of the abdominal aorta. It is usually asymptomatic while is with high mortality once rupture \cite{kent2014abdominal}. This rupture is usually avoided via placing a stent graft in the aneurysm area to re-establish the blood flow and hence to exclude the aneurysm wall from blood pressure. Fenestrated Endovascular Aortic Repair (FEVAR), which is specific for aneurysms near or include the renal and visceral vessels, uses fenestrated stent grafts with fenestrations, scallops and branch stent grafts to perfuse main and branch vessels \cite{cross2012fenestrated}. A real main fenestrated stent graft and its model are shown in \cref{fig: introduction}a and b respectively. One main challenge in FEVAR is the cannulation of a branch stent graft from the fenestration or scallop into the corresponding branch vessel. Robot-assisted systems have been developed to facilitate this challenge, i.e. the Magellan system (Hansen Medical, CA, USA). However, only 2D fluoroscopy images are currently used for navigation while the cannulation needs precise 3D-3D geometrical alignments.

Various methods have been developed by researchers to improve the navigation. The stent graft delivery device was tracked and detected by Frangi filtering and Robust Principal Component Analysis (RPCA) \cite{volpi2015online}. The 3D stent shape was recovered from one X-ray image by registration and semi-simultaneous optimization \cite{demirci20113d}. The aortic and illiac deformations caused by device insertions were corrected by the skeleton-based As-Rigid-As-Possible (ARAP) method \cite{toth2015adaption}. The 3D shape of a fenestrated stent graft after its deployment was instantiated semi-automatically from one fluoroscopy image of its compressed state with markers and the RP5P method \cite{zhoustent}. The 3D shape of a fenestrated and deployed stent graft was instantiated semi-automatically from one fluoroscopy image of its deployed state with the RP5P method, graft gap interpolation and semi-automatic marker center determination \cite{zhou2018real}. In \cite{zhou2018real}, markers could only be segmented into one class while manual classification was essential for 3D shape instantiation.

In this paper, an automatic 3D shape instantiation is possible, as the markers in \cite{zhou2018real} were segmented into multiple-classes automatically though being designed into five different shapes. One experimental fluoroscopy image with the customized markers labeled in different colors is shown in \cref{fig: introduction}c. We use marker segmentation rather than marker detection to determine the marker center position, as segmentation is a pixel-level classification and is more precise. There are two challenges in segmenting these customized markers into multiple-classes: 1) the markers are very small (the reason will be explained in \cref{Sec: Marker Design}), causing class-imbalance problems; 2) the markers are with similar appearances (the reason will be explained in \cref{Sec: Image Collection}).
 
Compared to conventional segmentation methods, deep convolutional neural network which extracts and classifies the features automatically with the using of multiple non-linear modules has outperformed significantly in semantic segmentation. Fully Convolutional Neural Network (FCNN) was the very first proposed network which improved the image-level classification with Convolutional Neural Network (CNN) to a pixel-level classification with the using of fully convolutional layers, deconvolutional layers and skip architectures \cite{long2015fully}. Ronneberger et al. firstly introduced FCNN into biomedical segmentation and proposed U-Net on neuronal structure segmentation and cell segmentation \cite{ronneberger2015u}. The Deeplab series including Deeplabv1 \cite{chen2014semantic}, Deeplabv2 \cite{chen2018deeplab}, Deeplabv3 \cite{chen2017rethinking}, and Deeplabv3+ \cite{chen2018encoder} with Atrous convolution, Atrous Spatial Pyramid Pooling (ASPP), and encoder-decoder modules were also popular networks in semantic segmentation.

Class-imbalance, where the background pixel number is much larger than the foreground pixel number, is a common challenging problem in semantic segmentation. Allocating large weights for the foreground pixels while allocating small weights for the background pixels were usually used to concentrate the training more on foreground pixels \cite{ronneberger2015u}. Three shortages exist when applying weighted loss in our application (will be proved in  \cref{Sec: character}): 1) the weight needs to be manually set; 2) when the weight is too small, weighted loss could not distinguish between different foreground classes, while if the weight is too large, the background would be mis-classified as a foreground; 3) its performance is insufficient.

Two-stage networks were also widely explored in both biomedical and natural community to improve the network performance on small object or class-imbalance segmentation. Cascade Fully Convolutional Network (CFCN) was proposed to segment the liver first as a Region of Interest (RoI), and then another FCN was trained to segment the small lesion inside the liver RoI \cite{christ2016automatic}. In Zhou et al.'s work, the pancreas was segmented firstly, and then the cyst inside the pancreas was segmented to improve the accuracy of the small cyst segmentation \cite{zhou2017deep}. In natural community, Mask Region-CNN (Mask R-CNN) was developed, where an object bounding box  was regressed and classified firstly and then FCN was applied inside this bounding box \cite{he2017mask}.

Apart from improving the network structure and using two-stage networks, various researches have also been carried out on the loss function. Topology aware FCN was proposed with considering multi-region topological relationships and smooth boundaries into the loss function for histology gland segmentation \cite{bentaieb2016topology}. Convolutional AutoEncoder (CAE) was added to the loss function to consider the shape prior for semantic segmentation, which shown improved results in the kidney ultrasound image segmentation \cite{ravishankar2017learning}. Recently, focal loss was introduced in the object detection domain, which added different scaling factors automatically to focus on training hard examples \cite{lin2017focal}. However, directly applying the focal loss in \cite{lin2017focal} into our application has three challenges: 1) the performance is insufficient (will be proved in  \cref{Sec: comparison}); 2) it needs careful parameter initialization; 3) the weight used in \cite{lin2017focal} would introduce the same problems as stated before for the weighted loss.

\begin{figure}[thpb]
\centering
\includegraphics[width=\linewidth]{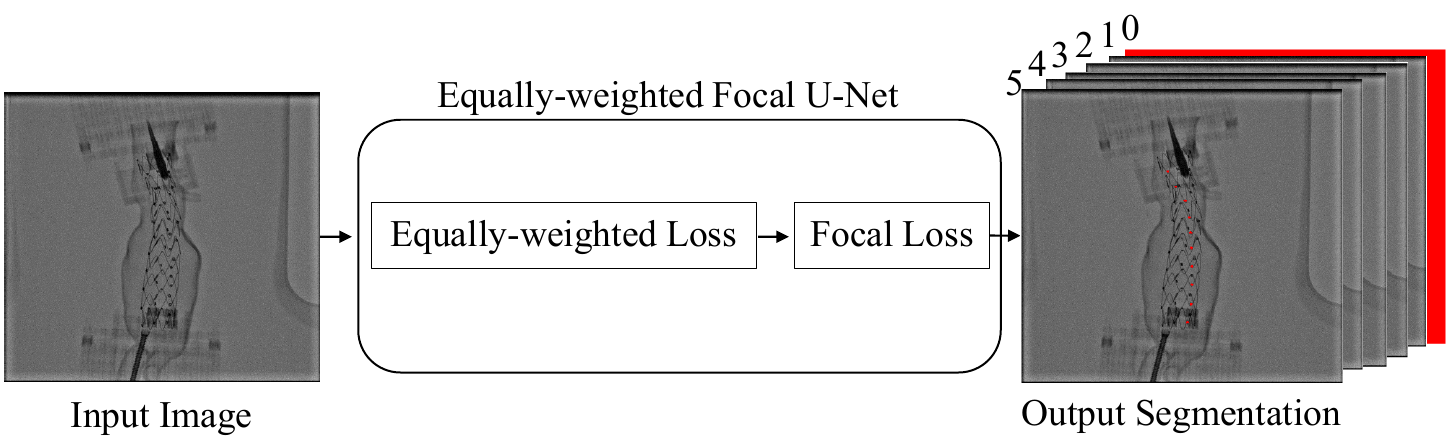}
\caption{The framework of the proposed Equally-weighted Focal U-Net: the output map is consisted of six classes: class $0$ represents the background, class $1-5$ represent the marker 1, marker 2, marker 3, marker 4 and marker 5. Red color indicates the pixels with probability of 1 in each output class.}
\label{fig: Workflow}
\end{figure}

In this paper, Equally-weighted Focal U-Net was proposed. "Equally-weighted" means equal weight of 1 was applied to the foreground and the background. "Focal" means focal loss was used. The proposed method is a one-stage network but with two-step training, as shown in \cref{fig: Workflow}. Firstly, U-Net with equally-weighted loss function was applied to segment a preliminary result. Secondly, U-Net with equally-weighted focal loss was used to improve the preliminary segmentation. It outperformed  the focal loss in \cite{lin2017focal} and Weighted U-Net in \cite{ronneberger2015u} in: 1) the model trained by equally-weighted loss is used as the initialization for later equally-weighted focal loss, avoiding careful manual parameter initialization; 2) equally-weighted loss avoids the possible problems caused by weighted loss and also reduces one hyper-parameter - the weight; 3) even though equally-weighted loss under-performs weighted loss, the later equally-weighted focal loss will improve the preliminary segmentation result and outperform weighted loss. U-Net was selected as the network structure, as it is easy to be trained from scratch with limited training data (80 images in this paper). The proposed Equally-weighted Focal U-Net and also the 3D shape instantiation were validated on 78 testing images, showing comparable results.

The \cref{Sec: method} describes the methodologies used in this paper, including marker design, image collection, Equally-weighted Focal U-Net, brief introduction of 3D shape instantiation, and experimental setup. In \cref{Sec: Results},  the impact of block number, data augmentation, and image enhancement are explored, the comparison between different methods is carried out, as well as the performance of segmentation and 3D shape instantiation are shown. The discussion and conclusion of the proposed method are summarized in \cref{Sec: Discussion} and \cref{Sec: conclusion} respectively.
 
\section{Methodology} \label{Sec: method}
The design of stent graft markers is described in \cref{Sec: Marker Design}. The \cref{Sec: Image Collection} introduces the progress of image collection. The \cref{Sec: Equally-weighted Focal U-Net} explains the data representation, deep learning structure and loss function used in the proposed Equally-weighted Focal U-Net. 3D shape instantiation \cite{zhou2018real} is briefly introduced in \cref{Sec: method instantiation} to facilitate understanding. In \cref{Sec: Experimental Setup}, parameters for experimental setup are described and explained.

\subsection{Marker Design} \label{Sec: Marker Design}
Stent graft markers were designed based on commercially-used gold markers (shown in \cref{fig: introduction}a) into five different shapes and were placed at five non-planar positions on each stent segment. The marker parameters are shown in \cref{tab:markers}. The lengths were designed to be similar to that of commercial markers which are around $1-3mm$. The thicknesses were empirically-determined for both minimized thickness and good imaging quality under lowest-radiation fluoroscopy. The shapes were designed with maximum differentiation and to be easily sewn onto the stents. Due to the high price of gold, these markers were printed on a Mlab Cusing R machine (ConceptLaser, Lichtenfels, Germany) with SS316L stainless steel powder for the experiment. The printed markers are shown in \cref{fig: introduction}d. The small marker size caused class-imbalance. The five marker classes occupied $0.03\%$, $0.01\%$, $0.02\%$, $0.03\%$, $0.03\%$ of the total pixels of the $512\times 512$ fluoroscopy image. 

\begin{table}
\centering
\caption{Marker Parameters}
\begin{tabular}{cccccc}
\hline
Marker Type & Circle & Sphere & Tube & Cross & Triangle \\
Marker Sequencing & 1 & 2 & 3 & 4 & 5 \\
\hline
Hole Radius (mm) & 0.5 & 0.2 & 0.2 &- & 0.63\\ 
Thickness (mm)       & 0.8 & 0.8 & 0.8 &0.8 & 0.8\\
Length (mm)      & 2.6 & -   & 2.5 &3 & 2.5\\
\hline
\end{tabular}
\label{tab:markers} 
\end{table}

\subsection{Image Collection}
\label{Sec: Image Collection}
For simulating the intra-operative fluoroscopy images in FEVAR, each stent segment of three stent grafts (illiac, fenestrated, and thoracic) was sewn with the five newly designed markers at non-planar positions, as shown in \cref{fig: introduction}c. The modified stent grafts were inserted, delivered and deployed into five 3D printed patient aneurysm phantoms. For more details of the 3D printed phantoms, please read \cite{zhou2018real}. Fourteen matching positions or setups were selected and each setup was scanned by a GE Innova 4100 (GE Healthcare, Bucks, UK) with 13 view angles from $-90^\circ$ to $90^\circ$. This varying view angle is necessary for proving that the 3D shape instantiation works for any view angle. It caused the 2D marker shape appearances to be similar in the fluoroscopy images, even though these markers were designed to be differentiable in 3D. During the experiment, one marker fell off which caused that setup to be abandoned. The operator forgot to store 11 fluoroscopy images, resulting 158 2D fluoroscopy images in total. $78/158$ images from 6 setups with complete 13 view angles were used for the testing while others were used for the training. Due to the limited number of available images, no evaluation images were split. More details about the experimental setup and image collection could be found in \cite{zhou2018real}.

\subsection{Equally-weighted Focal U-Net}
\label{Sec: Equally-weighted Focal U-Net}
\subsubsection{Data representation}
Given a training or testing data set $\{\textbf{I}_1,\textbf{I}_2,...,\textbf{I}_k,...,\textbf{I}_{\rm K}\}, k\in[1,\rm K]$, where $\textbf{I}_k$ is one image example with width $\rm W$ and height $\rm H$, $\rm W =\rm H=512$ in this paper, $\rm K$ is the total number of images in the training or testing data set. The intensity of each pixel in $\textbf{I}_k$ is normalized into $[0, 1]$ by: $\textbf{I}_{norm_k}=\frac{\textbf{I}_k-min(\textbf{I}_k)}{max(\textbf{I}_k)}$. The segmentation ground truth of $\textbf{I}_k$ in the training data set is labelled as a labelling cube: $\textbf{L}_k=\{\textbf{L}_{k0},\textbf{L}_{k1},...,\textbf{L}_{kn},...,\textbf{L}_{k \rm N}\}, n\in[0,\rm N]$, where $\rm N$ is the number of marker classes, $\rm N = 5$ in this paper (\cref{fig: Workflow}), $\textbf{L}_{kn}$ has the same width $\rm W$ and height $\rm H$, $\textbf{L}_{k0}$ is the background labelling layer with background pixels labelled as $1$ and other pixels labelled as $0$, $\textbf{L}_{kn}$ is the $n^{th}$ class foreground or marker labelling layer with the $n^{th}$ class marker pixels labelled as $1$ and other pixels labelled as $0$. Since the markers are very small, those markers do not fully overlap each other frequently during the varying fluoroscopy view angle. Hence, it is reasonable to consider the multiple-class marker segmentation as a no-overlap problem, where one pixel only belongs to one class.

\subsubsection{U-Net structure}
\begin{figure}[thpb]
\centering
\includegraphics[width=\linewidth]{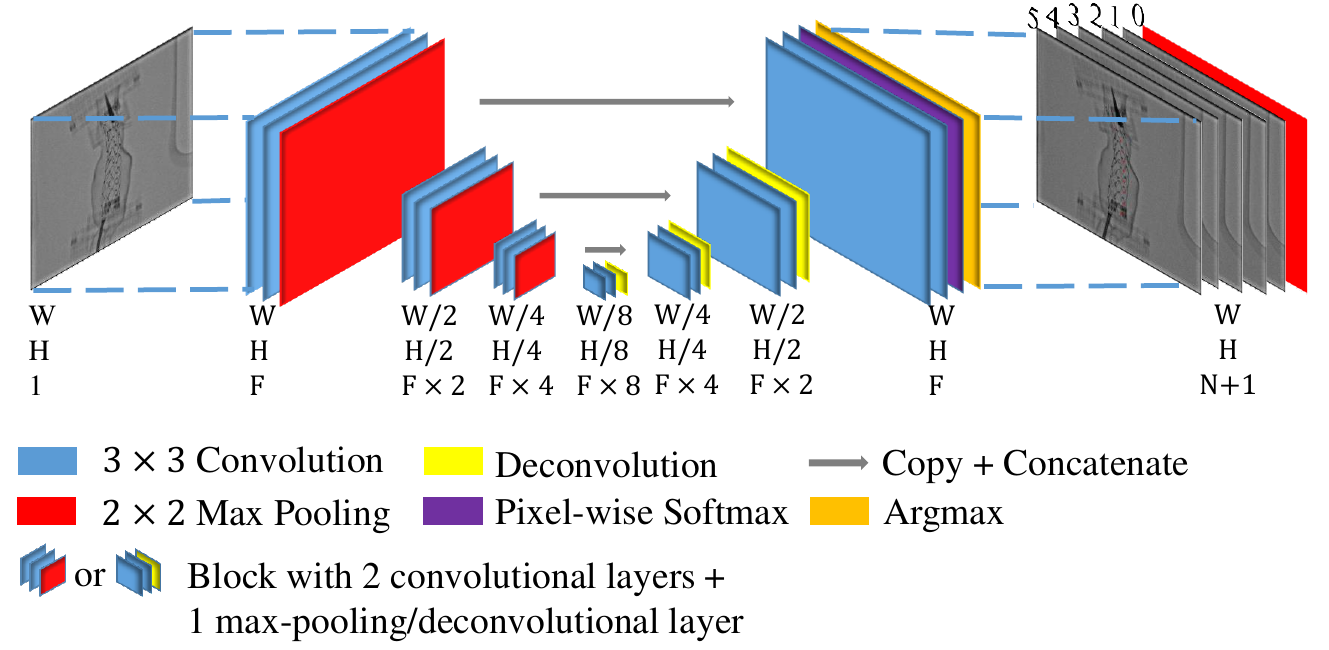}
\caption{An illustration of a 3-block U-Net: three max-pooling/deconvolutional layers are used in total, two convolutional layers are used in each block, the width W and height H of the image are half/twice while the number of feature channel (F) is twice/half after each max-pooling/deconvolutional layer, $N=5$ in this paper.}
\label{fig: U-Net}
\end{figure}

According to the U-net structure \cite{ronneberger2015u}, a normalized image $\textbf{I}_{norm_k}$ is passed into the proposed network as an input, then a probability map cube $\textbf{P}_k=\{\textbf{P}_{k0},\textbf{P}_{k1},...,\textbf{P}_{kn},...,\textbf{P}_{k\rm N}\}, n\in[0,\rm N]$ is calculated, where $\textbf{P}_{kn}$ is with the same width $\rm W$ and height $\rm H$. The value of each pixel in $\textbf{P}_{kn}$ is the probability of that pixel belongs to the $n^{th}$ class and is between $[0, 1]$. The network structure used in this paper is consisted of convolutional layers, max-pooling layers and deconvolutional layers, as illustrated in \cref{fig: U-Net}. It has two paths: a contracting path (left) and an expansive path (right). For convenience, we term the layers that manipulate on images with the same size as a block. In the contracting path, each block is consisted of two convolutional layers following by a max-pooling layer. In the expansive path, each block is consisted of two convolutional layers following by a deconvolutional layer. The last block is consisted of two convolutional layers, a $1\times 1$ convolutional layer, a pixel-wise softmax layer, and an argmax layer. The network in \cref{fig: U-Net} is defined as a 3-block U-Net, as three max-pooling/deconvolutional layers are used in total. In this paper, the stride for the convolutional layer is always $1$ while that for the max-pooling layer is always $2$.

\subsubsection{Loss function}
After passing $\textbf{I}_{norm_k}$ through the U-Net, each pixel will have a U-Net-predicted value for the $\rm N+1$ classes: $y_0,y_1,...,y_n,...,y_{\rm N},n\in[0,\rm N]$. Pixel-wise softmax is used to transform $y_{n}$ into the probability $p_n\in [0, 1]$ by:
\begin{equation}
  p_{n}=\frac{e^{y_{n}}}{\sum_{i=0}^{\rm N} e^{y_{i}}}\\
\label{equ:ce_loss}
\end{equation}
Cross-entropy loss is calculated across the labelling and predicted probability cube to measure the difference between the predicted probability $\textbf{P}$ and the ground truth $\textbf{L}$:
\begin{equation}
CE_{loss} = -\sum_{i=1}^{\rm W} \sum_{j=1}^{\rm H} \sum_{n=0}^{\rm N}\textbf{L}_{(i,j,n)}log(\textbf{P}_{(i,j,n)})
\label{equ:weight_ce_loss}
\end{equation}
Usually, weighted loss was applied to solve the class-imbalance problem: 
\begin{equation}
WCE_{loss} = -\sum_{i=1}^{\rm W} \sum_{j=1}^{\rm H} \sum_{n=0}^{\rm N}W_n\textbf{L}_{(i,j,n)}log(\textbf{P}_{(i,j,n)})
\label{equ:weight_ce_loss}
\end{equation}
Here, $W_0=1$ while $W_n>1, n\in [1, \rm N]$. In this paper, equally-weighted loss was applied for the first-step training. $W_n=1, n\in [0, \rm N]$. When the loss converges to a minimum, equally-weighted focal loss was applied to improve the preliminary segmentation results:

\begin{equation}
Focal_{loss} = - \sum_{i=1}^{\rm W} \sum_{j=1}^{\rm H} \sum_{n=0}^{\rm N}(1-\textbf{P}_{(i,j,n)})^2\textbf{L}_{(i,j,n)}log(\textbf{P}_{(i,j,n)})
\label{equ:focal_ce_loss}
\end{equation}

The scaling factor of $(1-\textbf{P}_{(i,j,n)})^2$ suppresses heavily the loss contribution of correctly-segmented pixels (when $\textbf{P}_{(i,j,n)}=0.9,(1-\textbf{P}_{(i,j,n)})^2 =0.01$). However, it suppresses lightly the loss contribution of wrongly-segmented pixels (when $\textbf{P}_{(i,j,n)}=0.1,(1-\textbf{P}_{(i,j,n)})^2 =0.81$). Thus the focal loss concentrates the training on wrongly-segmented pixels or hard pixels.

\subsection{3D Shape Instantiation}
\label{Sec: method instantiation}
The marker center positions segmented by the proposed Equally-weighted Focal U-Net were used as the input for the RP5P method to recover the 3D pose of each stent segment. The whole stent graft shape was then recovered by graft gap interpolation. Details of the 3D shape instantiation could be found in \cite{zhou2018real}. Its codes are also available on-line.

\subsection{Experimental Setup}
\label{Sec: Experimental Setup}
\subsubsection{Data augmentation} to evaluate the character of the proposed network to data augmentation, two different data augmentation methods were compared: 1) rotated the $80$ training images from $-36^\circ$ to $35^\circ$ with $1^\circ$ as the interval; 2) rotated the $80$ training images from $-180^\circ$ to $165^\circ$ with $15^\circ$ as the interval and flipped each rotated image along the horizontal and vertical direction respectively. Both data augmentation methods augmented the training images with $72$ times, resulting $5760$ training images.

\subsubsection{Image enhancement} to evaluate the performance of the proposed network to image enhancement, image intensity adjustment and contrast-limited adaptive histogram equalization were applied with $MATLAB ^{ \textcopyright} $ function:
\begin{equation}
\textbf{I}'_k = adapthisteq(imadjust(\textbf{I}_{norm_k}))
\end{equation}

\subsubsection{Ground truth labelling} the markers were labelled in Analyze (AnalyzeDirect Inc, Overland Park, KS, USA) with firstly magnifying the image from $512\times 512$ to $4096\times 4096$ and then shrinking the image from $4096\times 4096$ back to $512\times 512$. Hence, the 1 pixel error of labelling in the $4096\times 4096$ resolution image would be shrunk to $0.125$ pixel in the $512\times 512$ resolution image.

\textit{Other parameters:} the learning rate was set step-wisely and divided by two or five when the loss stopped decreasing. The dropout rate was set as 0.75. The weights in the neural network were initialized by truncated normal distribution with $mean=0.0$ and $std=0.1$ while the biases were initialized by constant $0.1$. The optimizer was the momentum optimizer in Tensorflow with the momentum set as 0.95. The batch size was set as 1. The loss function was written by tf.nn.softmax, tf.log, and tf.reduce\_mean, which may present slightly worse stability than that using default tf.nn.softmax\_cross\_entropy\_with\_logits. The mean Intersection over Union (mIoU), the overlap of the ground truth and the prediction over the union of the ground truth and the prediction, was calculated to evaluate the segmentation performance. Except \cref{Sec: character}, all training procedures were based on the data augmented with $30^\circ$ image rotation and without image enhancement.

\section{Results}
\label{Sec: Results}
The characters of the proposed network with respect to the number of U-Net block, image enhancement, data augmentation, and weight are illustrated in  \cref{Sec: character}. The comparison between different methods is presented in  \cref{Sec: comparison}. Detailed multiple-class marker segmentation results are shown in \cref{Sec: segmentation}. The accuracy of 3D shape instantiation based on the marker segmentation in this paper is presented in \cref{Sec: instantiation}.

\subsection{Network Characters}
\label{Sec: character}
The mIoUs achieved with different setups are shown in \cref{tab: impact}, where the highest mIoU is emphasized in bold font.
\begin{table*}
\centering
\caption{U-Net with different setups (mIoU-mean Intersection over Union, B-Background, M-Marker, Num.-Number)}
\begin{tabular}{ccccccccccccc}
\hline
Row&$30^\circ$&$180^\circ$&Image&Block &Weight&Focal&B&M1&M2&M3&M4&M5\\
&Augmentation&Augmentation&Enhancement&Num.&&Loss&mIoU&
mIoU&mIoU&mIoU&mIoU&mIoU\\
\hline
1&$\surd$&&&1&1&$\surd$&\textbf{0.9996}&0.6392&0.5159&0.5929&\textbf{0.5692}&0.5998\\

2&$\surd$&&&2&1&$\surd$&\textbf{0.9996}&0.7030&0.5687&0.6778&0.5094&0.5765\\

3&$\surd$&&&3&1&$\surd$&\textbf{0.9996}&\textbf{0.7325}&\textbf{0.5828}&0.6952&0.5453&\textbf{0.6105}\\

4&$\surd$&&&4&1&$\surd$&\textbf{0.9996}&0.7280&0.5462&0.6831&0.5266&0.5883\\

5&$\surd$&&&5&1&$\surd$&\textbf{0.9996}&0.7254&0.5395&0.6843&0.5156&0.5841\\

6&$\surd$&&&6&1&$\surd$&\textbf{0.9996}&0.6179&0.5475&0.5596&0.4424&0.4986\\

7&&$\surd$&&2&1&$\surd$&\textbf{0.9996}&0.4793&0.5081&\textbf{0.6988}&0.5523&0.5001\\

8&$\surd$&&$\surd$&2&1&$\surd$&0.9992&0.4092&0.0843&0.2779&0.0993&0.3133\\

9&$\surd$&&&2&1&&0.9993&0.1900&0.0000&0.0000&0.0000&0.0000\\

10&$\surd$&&&2&20&&0.9986&0.1508&0.0428&0.1151&0.1311&0.1387\\

11&$\surd$&&&2&50&&0.9981&0.4168&0.2260&0.3639&0.3037&0.3630\\

12&$\surd$&&&2&100&&0.9979&0.4222&0.2195&0.3439&0.2978&0.3415\\

13&$\surd$&&&2&500&&0.9967&0.3020&0.1207&0.2782&0.2531&0.2868\\
\hline
\end{tabular}
\label{tab: impact} 
\end{table*}

\subsubsection{Number of U-Net block} \label{Sec: layer}
Equally-weighted Focal U-Net with block number from $1-6$ were trained to segment the multiple-class markers in \cref{fig: Workflow}, mIoUs are listed in Row $1-6$ in \cref{tab: impact}. It can be concluded that 1-block U-Net and 6-block U-Net under-performed slightly others. However, the training time increased from 36 hours for 1-block U-Net to 120 hours for 6-block U-Net. Based on this comparison result, 2-block U-Net was chosen as a trade-off between the efficiency and the performance in the following validations. 

\subsubsection{Data augmentation} \label{Sec: data_augmentation}
Equally-weighted Focal U-Net with 2 blocks was trained on the data augmented with $30^\circ$ image rotation and with $180^\circ$ image rotation respectively. The mIoUs for six classes on the 78 testing images are summarized in the Row 2 and the Row 7 in \cref{tab: impact}. The results showed that the mIoUs achieved with $30^\circ$ image rotation are higher than that with $180^\circ$ image rotation in most classes, except for Marker 3 and Marker 4. Hence, $30^\circ$ image rotation was utilized as data augmentation in this paper.

\subsubsection{Image enhancement} \label{Sec: image_enhancement}
Equally-weighted Focal U-Net with 2 blocks was trained on the training data with and without image enhancement respectively. The mIoUs of the six classes achieved on the 78 testing images are summarized in the Row 2 and the Row 8 in \cref{tab: impact}. The results presented that the mIoUs decreased significantly when the training data was pre-processed with image enhancement. Therefore, the images in the training set will only be processed by normalization in the following training.

\subsubsection{Weight} \label{Sec: weight}
2-block U-Net with the weight of 1, 20, 50, 100, 500 were trained respectively. The mIoUs of the six classes on the 78 testing images are listed in the Row 9-13 in \cref{tab: impact}. The results illustrated that 2-block U-net with the weight of $50$ presented optimal performance comparing with small weights (weight = 1, 20) and the large weight (weight = 500). Thus, 2-block U-Net with the weight of 50 was applied in the following work.

\begin{figure}[thpb]
\centering
\includegraphics[width=\linewidth]{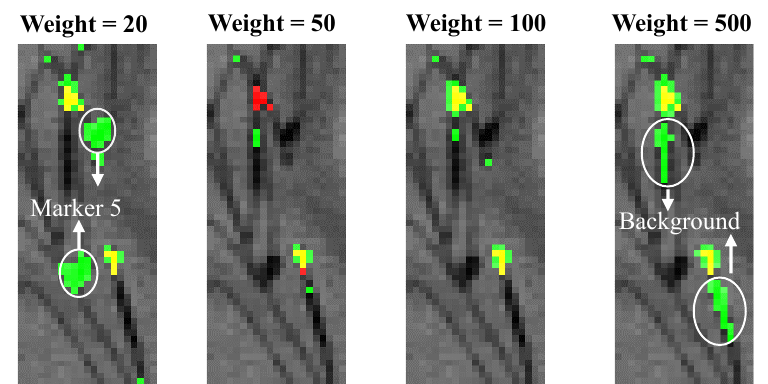}
\caption{Cropped segmentation results for Marker 2 with the weight as 20, 50, 100, and 500, where red region - the ground truth, green region - the prediction, yellow region -  the correctly-segmented pixels.} 
\label{fig: weight_illustration}
\end{figure}

The segmentation results of the 2-block U-Net with different weights are illustrated in \cref{fig: weight_illustration}. It can be seen that the five foreground or marker classes could not be clearly distinguished between each other with a small weight, i.e. $20$. However, if the weight of the network is too large, i.e. $500$, the background was mis-classified as a foreground, as this wrong classification contributed too less to the total loss. For example, a wrongly-segmented background ($\textbf{P}_{(i,j,n)}=0.1$) contributed $(1-\textbf{P}_{(i,j,n)})\times 1 =0.9$ to the total loss while a wrongly-segmented foreground ($\textbf{P}_{(i,j,n)}=0.1$) contributed $(1-\textbf{P}_{(i,j,n)})\times 500 =450$ to the total loss. The mIoUs of the background decreased along the increased weight (Row 9-13 in \cref{tab: impact}), which also proves this trend.

\subsection{Comparison between different methods}
\label{Sec: comparison}
\begin{figure}[thpb]
\centering
\includegraphics[width=\linewidth]{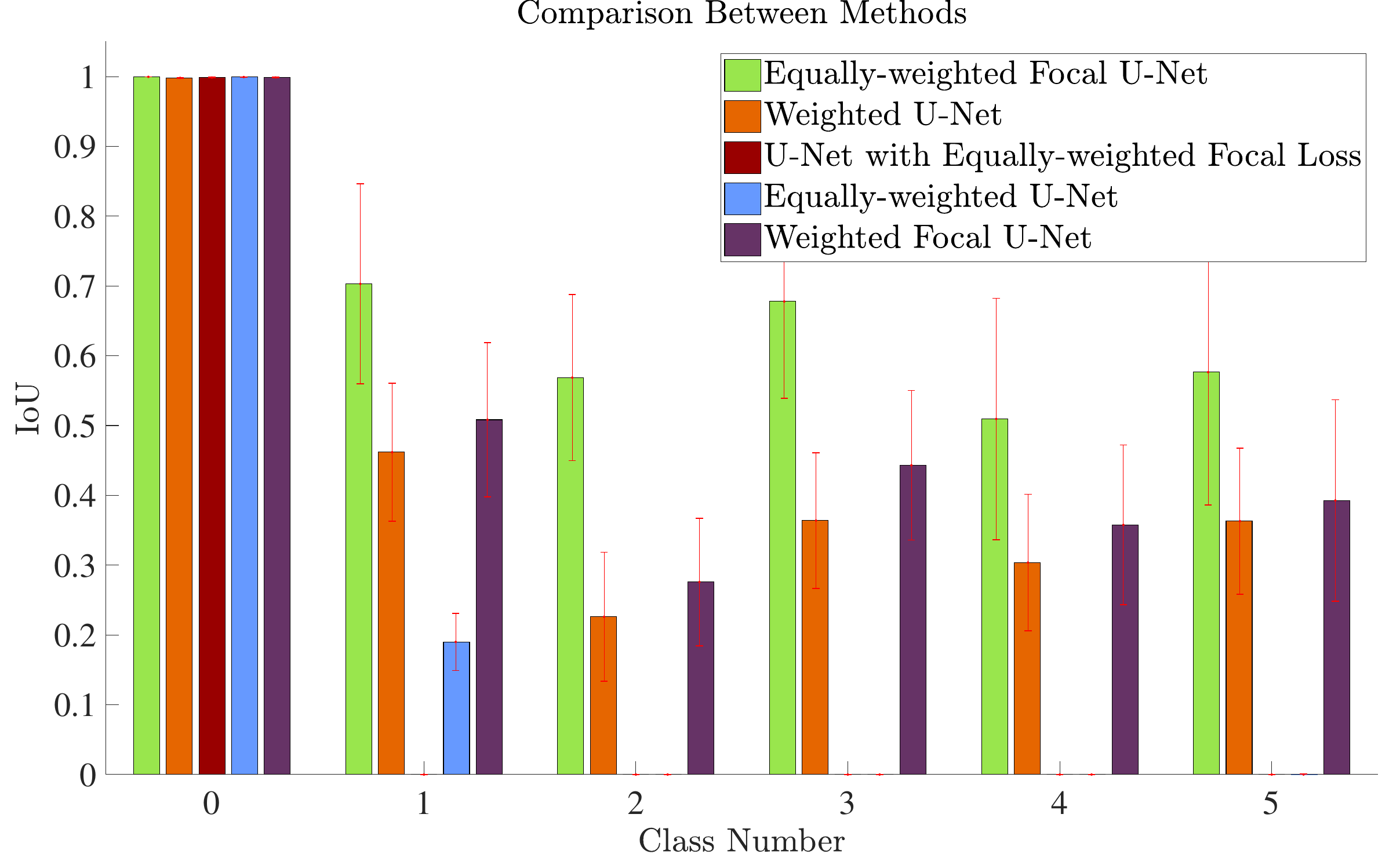}
\caption{The $mean \pm std$ IoUs for the six classes segmented by five different methods} 
\label{fig: comparison}
\end{figure}

The performance of 2-block U-Net using five different methods were explored in \cref{fig: comparison}: 1) Equally-weighted Focal U-Net (the proposed method); 2) Weighted U-Net with the weight as 50 for foreground and the weight as 1 for background; 3) U-Net with Equally-weighted Focal Loss which used an equally-weighted focal loss from the beginning of the training; 4) Equally-weighted U-Net with the weight set as 1 for both the foreground and the background; 5) Weighted Focal U-Net with the weight set as 50 for the first step training, and then focal loss with the weight of 50 for the second step training. The performance of these methods are shown by the mean and std IoUs. The \cref{fig: comparison} illustrated that the proposed method has outstanding performance on every marker class comparing with other methods.

\subsection{Multiple-class Marker Segmentation}
\label{Sec: segmentation}
\begin{figure}[thpb]
\centering
\includegraphics[width=\linewidth]{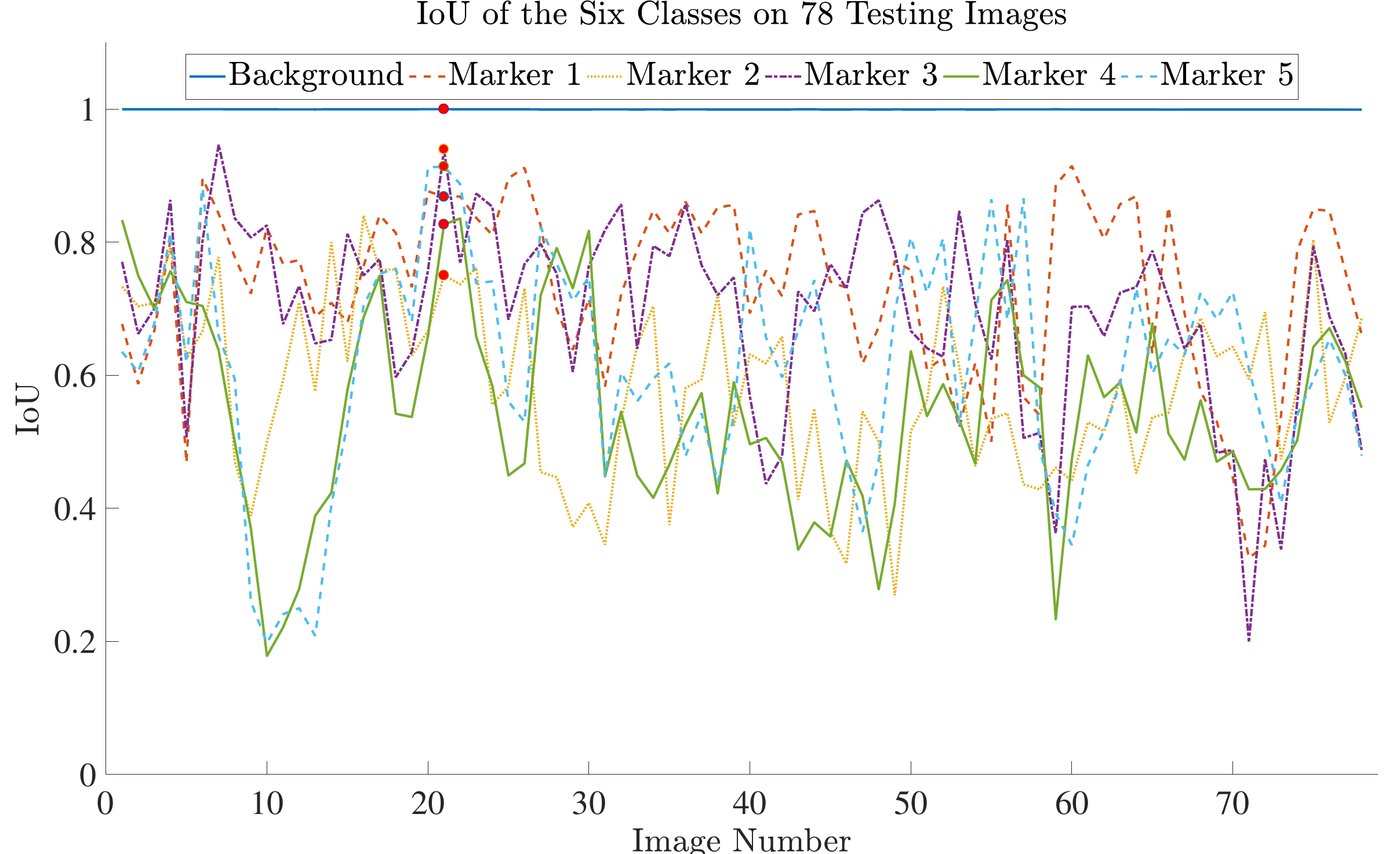}
\caption{The IoU of the six classes on 78 testing images segmented with a 4-block Equally-weighted Focal U-Net.} 
\label{fig: IoU}
\end{figure}

\begin{figure}[thpb]
\centering
\includegraphics[width=\linewidth]{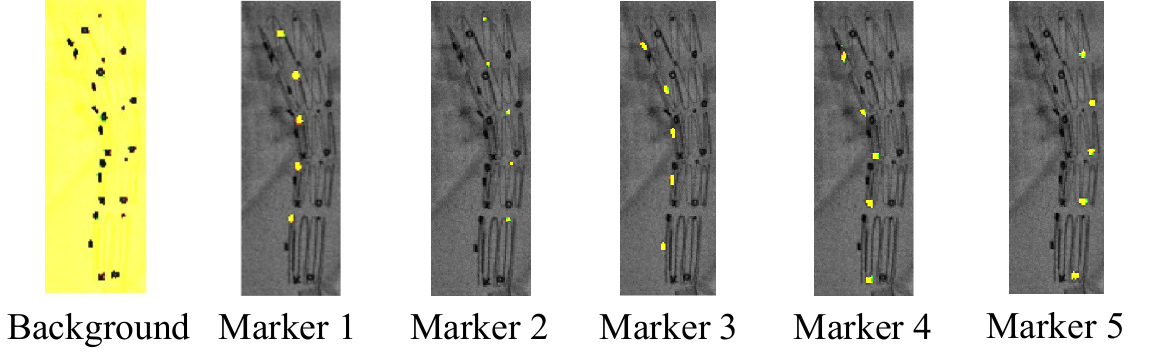}
\caption{Cropped segmentation results for six classes on image NO.21: red - the ground truth, green - the prediction, yellow -  the correctly-segmented pixels.} 
\label{fig: segmentation}
\end{figure}

Equally-weighted Focal U-Net with 3-block (Row 3 in \cref{tab: impact}) was applied to segment each testing image. The results are illustrated in \cref{fig: IoU}. The \cref{fig: IoU} showed that the proposed network could segment most of the images with outstanding performance, except from a few markers in the image No.10, No.13, No.59 and No.71. Besides, the \cref{fig: segmentation} presents the segmentation details of image No.21 using the proposed method, where each marker class was segmented with a high overlap between the ground truth and the prediction.

\subsection{3D Shape Instantiation}
\label{Sec: instantiation}
\begin{figure}[thpb]
\centering
\includegraphics[width=\linewidth]{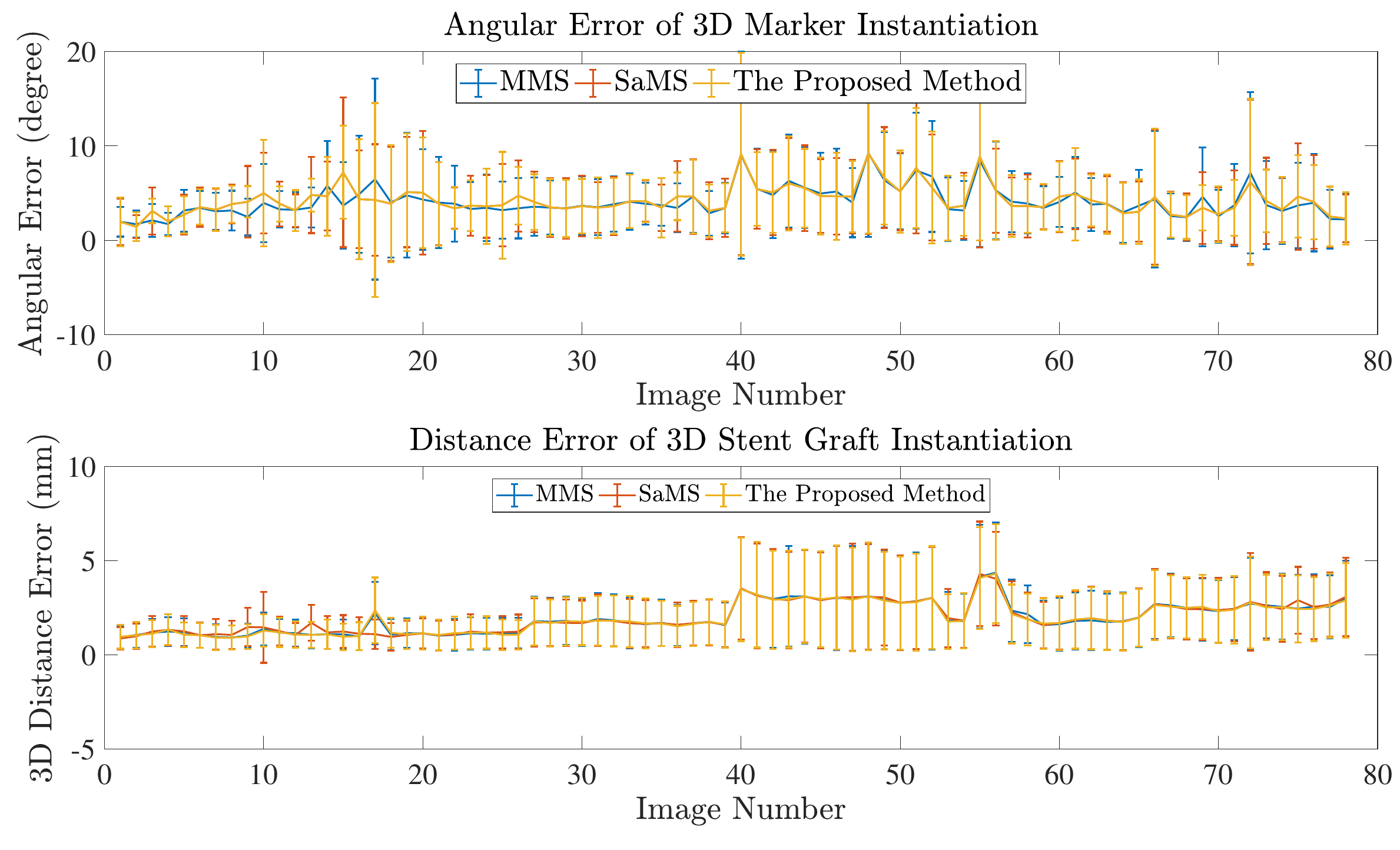}
\caption{The angular error of 3D marker instantiation (top) and the 3D distance error of 3D stent graft instantiation (bottom) using three different marker center determination methods: MMS - Manual Marker Segmentation; SaMS - Semi-automatic Marker Segmentation; the proposed method in this paper} 
\label{fig: instantiation}
\end{figure}

The 78 images contain 2470 markers, $81.01\%$ of them were segmented with a center position error $<1.6mm$ which are 2 pixels on the fluoroscopy image. The marker center positions determined with $>1.6mm$ error were corrected manually. With these marker center positions, the angular error and 3D distance error of 3D shape instantiation were illustrated in \cref{fig: instantiation}, showing that the proposed method presents comparable performance with 3D shape instantiation with both manual and semi-automatic marker center determination. More 3D shape instantiation results could be found in \cite{zhou2018real}.

All the training procedures were based on a NVIDIA TITAN Xp GPU. Segmenting one image took less than $0.1s$. The programming is based on the released code of \cite{akeret2017radio}.

\section{Discussion}
\label{Sec: Discussion}
Equally-weighted Focal U-Net was proposed to segment the customized stent graft markers into multiple-classes. The segmented marker center positions would be used by the RP5P method and hence automatic 3D stent graft shape instantiation was possibly achieved. Focal loss was successfully applied into semantic segmentation for the first time with convincing improvements.

In \cref{Sec: layer},the performance of U-Net with different block number was explored. The results showed that Equally-weighted Focal U-Net did not achieve higher mIoU along with an increasing block number. One possible reason could be network degradation. In the future, the network structure will be explored in details.

In \cref{Sec: weight}, different weights were explored. Usually, weighted loss outperforms equally-weighted loss for class-imbalance segmentation, as it treats the foreground more importantly by assigning a higher weight for it. However, in this paper, we consider the background as equally important as the foreground, as a mis-classified background will also decrease the foreground IoU. So equally-weighted loss was applied. 

The proposed method is capable for multiple-class marker segmentation, obtained an overall mIoU of 0.6943, and detected $81.01\%$ markers with center position error $<1.6mm$. Comparable 3D shape instantiation error was achieved ($1.9605mm$) with the approximately-automatic marker center determination method in this paper, with respect to 3D shape instantiation with semi-automatic marker center determination ($1.9746mm$) and with manual marker center determination ($1.9874mm$) in \cite{zhou2018real}.

\section{Conclusions}
\label{Sec: conclusion}
In this paper, Equally-weighted Focal U-Net was proposed for multiple-class marker segmentation and then automatic 3D stent graft shape instantiation could be achieved. The performance of the proposed network was explored and discussed with different characters, such as the number of blocks, method of data augmentation, image enhancement, and different weights. Based on these results, 3-block Equally-weighted Focal U-Net showed optimal accuracy in multiple-class marker segmentation. In the future, the proposed network will be further improved and extended to a general framework for wider applications.
 
\section*{ACKNOWLEDGMENT}
We gratefully acknowledge the support of NVIDIA Corporation with the donation of the Titan Xp GPU used for this research. This work was supported by EPSRC project grant EP/L020688/1.

\bibliographystyle{IEEEtran}
\bibliography{IROS2018}
\end{document}